\documentclass[10pt,twocolumn,letterpaper]{article}

\usepackage{iccv}
\usepackage{times}
\usepackage{epsfig}
\usepackage{graphicx}
\usepackage{amsmath}
\usepackage{amssymb}

\usepackage{booktabs}

\usepackage{xspace}
\usepackage{multirow}
\usepackage{siunitx}
\usepackage{amsmath}
\usepackage{colortbl}
\usepackage{soul}

\usepackage[breaklinks=true,bookmarks=false]{hyperref}

\usepackage{authblk}

\DeclareMathOperator*{\argmax}{arg\,max}
\DeclareMathOperator*{\argmin}{arg\,min}

\usepackage[capitalize]{cleveref}
\crefname{section}{Sec.}{Secs.}
\Crefname{section}{Section}{Sections}
\Crefname{table}{Table}{Tables}
\crefname{table}{Tab.}{Tabs.}

\definecolor{darkblue}{rgb}{0.0, 0.0, 0.5}

\newcommand{\grood}{GROOD\xspace}
\newcommand{\na}{{\bf ---}\xspace}
\newcommand{\del}[1]{}
\newcommand{\first}[1]{\textcolor{darkblue}{\bf #1}}

\newcommand{\pz}{\phantom{0}}

\iccvfinalcopy 

\def\iccvPaperID{34} 

\ificcvfinal\fi

\begin{document}

\title{Calibrated Out-of-Distribution Detection with a Generic Representation}

\author[$\dagger$]{Tomáš Vojíř\thanks{Corresponding author, \tt vojirtom@fel.cvut.cz}}
\author[$\dagger$]{Jan Šochman}
\author[$\ddagger$]{Rahaf Aljundi}
\author[$\dagger$]{Jiří Matas}
\affil[$\dagger$]{\small CMP Visual Recognition Group, FEE, Czech Technical University in Prague}
\affil[$\ddagger$]{\small Toyota Motor Europe, Brussels, Belgium}

\maketitle
\ificcvfinal\thispagestyle{empty}\fi

\begin{abstract}
Out-of-distribution detection is a common issue in deploying vision models in
practice and solving it is an essential building block in safety critical
applications. Most of the existing OOD detection solutions focus on improving the OOD
robustness of a classification model trained \texttt{exclusively} on
in-distribution (ID) data.  In this work, we take a different approach and
propose to leverage generic pre-trained representation.
We propose a novel OOD method, called \grood, that formulates the OOD detection as a Neyman-Pearson task with well calibrated scores and which
achieves excellent performance, predicated by the use of a good generic
representation. 
Only a trivial training process is required for adapting \grood to
a particular problem. The method is simple, general, efficient, calibrated and
with only a few hyper-parameters. The method achieves state-of-the-art
performance on a number of OOD benchmarks, reaching near perfect
performance on several of them. 
The source code is available at \url{https://github.com/vojirt/GROOD}.
\end{abstract}


\section{Introduction}

The problem of detection of out-of-distribution data points, OOD in short, is
important in many computer vision
applications~\cite{Bergmann2021,Kuo2019,chan2021_bench}. One can even argue
that no model obtained by machine learning on a training set $\mathcal{T}$
should be deployed  without the  OOD ability, since in practice it is almost
never the case that all the models input data will be
drawn from the same distribution that generated~$\mathcal{T}$~\cite{Zhao2021}.
For undetected out-of-distribution data, the prediction will in general be
arbitrary, with possibly grave real-world consequences, especially in
safety-critical applications. 
The importance and ubiquity of OOD is evidenced
by the fact that virtually the same problem has emerged in different contexts
under different names - open set recognition, anomaly or outlier detection, and
one-class classification. 

The reasons for test data not being from the training set distribution are
diverse; they often influence the terminology used. In open set
recognition (OSR)~\cite{Mahdavi2021,yang2021generalized}, the semantic shift is
considered, \ie the introduction of new classes at test time. Failures of the
measurements system generate outlier data. In anomaly detection, the presence
of out-of-distribution data is assumed rare. A domain shift, e.g. when
a classifier trained on real-world images is applied to clip art, leads to
a severe data distribution change.

So far, prior art has mainly developed OOD detection models by supervised
training on in-distribution (ID)
data~\cite{yang2021generalized,Bogdoll2022,Bitterwolf2022}. We follow the recent success of 
self-supervised representation  model training~\cite{Radenovic_2023_CVPR, radford2021learning}, 
we apply it to out-of-distribution detection, our approach produces a calibrated 
decision strategy and we analyze its performance in various scenarios.

The performance of the proposed method is predicated by the use of a \textit{good generic representation}. 
Any {\it good} representation should enable solving a given, a priori unknown,
downstream task. A good {\it generic} representation should enable solving
multiple tasks without the need of fine-tuning on the task data. To verify the
goodness and generality of tested representations, we first exploit two commonly used
simple classifiers: (i) linear probe (LP), and (ii) the nearest mean (NM)
classifier. These simple classifiers already
outperform the state-of-the-art on a broad range of OOD detection problems, often by a large
margin, however, without apriori knowledge about the specificity of the OOD data its 
unclear which of these simple methods (or any other score based methods build on top of 
generic representation) should be preferred.

Since the LP and NM methods perform each well on different classes of the OOD
problems, we formulate a Neyman-Pearson task~\cite{Schlesinger2002,
Neyman1928, Neyman1933} on their combination. We call this approach \grood
(for Generic Representation based OOD detection). It models the in-distribution (ID)
as a 2D Gaussian in the space of LP and NM responses and provides a robust
solution to the OOD problem. It also naturally results in well calibrated rejection scores, which
allow us to define a global threshold for data rejection, \ie OOD identification. The global threshold 
can be set to incur user-specified pre-defined error on ID data and is calibrated for 
all classes, meaning, the pre-defined error is the same for all classes.
In contrast, most current state-of-the-art methods work on basis of similarity scores with no simple 
mechanism for selecting a single threshold for OOD rejection.
This novel approach significantly improves OOD performance and the experiments also confirm
the superiority of using the generic representation over problem-specific approaches
that train or fine-tune the feature extractor on a particular ID training set. 

The \grood does not require any information about the out-of-distribution data, \eg in the form of a few examples of the anomalies, and is
thus applicable to all the standard setting of the OOD and OSR
problems~{\cite{yang2021generalized}}. To summarize, the contributions of the paper are:

\begin{itemize} 
\item We show that using a generic pre-trained representation together with
    a simple classifier achieves state-of-the-art performance on a number of
        OOD benchmarks.

\item We formulate the OOD detection as a Neyman-Pearson task in the space of LP and NM
    scores. The operating point is selected by the allowed false negative rate
        for {\it all} ID classes. This results in a well calibrated
        classification score on the ID task.

\item We evaluate the method on a wide range of OOD problems. The proposed method outperforms the state-of-the-art by a large margin on
most of the problems and even saturates several commonly used benchmarks.

\end{itemize}

\section{Related Work}\label{sec:relatedwork}
Out-of-distribution (OOD) detection refers to the identification of test
samples  that are drawn from a different distribution than the underlying
training distribution of a given classification model. Hendrycks et
al.~\cite{hendrycks2016baseline} was one of the first to explore this problem
with modern neural networks using maximum softmax probability (MSP) obtained
from a classification model as a detection score. While being an classical
baseline in OOD detection, MSP can output high ID probabilities for unknown OOD
samples~\cite{sun2021react}. Subsequent work has attempted to provide more
robust OOD detection by either operating on a fixed model, or performing
additional ID training or even leveraging auxiliary OOD data. We refer an interested reader
to~\cite{yang2021generalized} for a complete survey on the different lines of
OOD detection research and cover only the main directions in this section.

\textit{Post-hoc methods} consider a pre-trained ID classifier and define different OOD detection scores or
perform manipulation of the input samples to increase the separability between
the distributions of ID and OOD
scores~\cite{liang2017enhancing,hsu2020generalized}. As a more robust
alternative to the MSP score, ~\cite{liu2020energy} proposed to use the energy
of the output logits as a scoring function showing strong improvements over MSP
and more separable scores. Later, \cite{hendrycks2019scaling} showed that
using maximum logit as an OOD detection score is significantly more robust than
MSP, suggesting that the normalization of the probability of the closed set
classes is the source for the overconfident predictions.

Of the distance based detection scores, we mention~\cite{lee2018simple} which
estimates the Mahalanobis distance to the closest class. Based on the
estimated $L_2$ distances in the learned embedding space, ~\cite{sun2022out}
propose instead to use the K-nearest neighbour (KNN) distance as a detection score.
This improves significantly over the Mahalanobis
distance. Manipulating the logits of a pre-trained ID
classifier has its limits though, which led to the second group of
approaches. 

\textit{Training based methods} target a stronger OOD detection performance
through regularizing the training such that the resulting classifier or
representation behave differently for ID compared to OOD inputs. Tackling the
same overconfident issue as in post-hoc methods,~\cite{wei2022mitigating}
proposed to train the ID classification model while enforcing a constant logit
norm. Deep ensemble~\cite{lakshminarayanan2017simple} combines adversarial
training with neural networks ensemble in addition to using the loss function
as a scoring rule. The computational cost of such approach might be prohibitive
for big networks. 

Other work aims at regularizing the training with virtual representatives of
OOD input. CSI~\cite{tack2020csi} utilizes contrastive training and apply
strong augmentations to the input images as an alternative to OOD data.
Adversarial Reciprocal Points Learning (ARPL)~\cite{Chen2021} proposes the
concept of "reciprocal" points as a proxy for OOD samples which are obtained by
combined discriminative and metric learning. This method showed state-of-the-art OOD detection performance, however, it requires complex training scheme and
large hyper-parameter tuning.  The proposed method is significantly more
efficient and improve over ARPL with a large margin in multiple benchmarks.

``A closed set classifier is all you need''~\cite{Vaze2022} suggests that an improved training scheme that leads to better performance on
ID data discrimination offers competitive OOD detection quality that
rivals that of OOD regularized training such as ARPL~\cite{Chen2021}. Further,~\cite{yangopenood} has evaluated a wide range of OOD detection
methods and their empirical results suggest that strong input augmentation
techniques, \eg MixUp~\cite{thulasidasan2019mixup},
CutMix~\cite{yun2019cutmix} and PixMix~\cite{hendrycks2022pixmix} are the most
effective type of training methods for OOD detection. 

Recently, the focus has shifted to OOD detection based on self-supervised pre-trained representations. A rotation prediction head was used in~\cite{hendrycks2019usingself}, while \cite{hendrycks2019usingpre} employs adversarial pre-training and label corruption, but still needs full network fine-tuning on the specific ID task.
SSD~\cite{sehwag2021ssd} first trains a feature extractor using contrastive self-supervised learning and then uses Mahalanobis distance to the class representants found by k-means.
MCM~\cite{ming2022delving} builds on CLIP~\cite{radford2021learning} by measuring a distance to "this is a photo of a \ldots" text encoder class prototype. However, they evaluate only on ImageNet-1k as ID, which was shown to be problematic~\cite{bitterwolf2023ninco}. Moreover, the text prompt prototypes in this form are limited to tasks with clearly defined objects.
ZOC~\cite{esmaeilpour2022zero} goes further in using the CLIP text encoder by dynamically generating semantically meaningful textual labels for each image and forming the ID score as a probability of classifying it into the seen classes instead of the generated ones. 

We follow the last group of methods by using a rich representation trained in self-supervised manner. However, we avoid using the text encoder, build the OOD detector in formally well defined way and evaluate on a wide and diverse range of OOD problems common in literature.

\section{The \grood Method}\label{sec:method}

\begin{figure*}[t]
    \centering
    \begin{tabular}{ccc}
	    \includegraphics[width=0.3\linewidth]{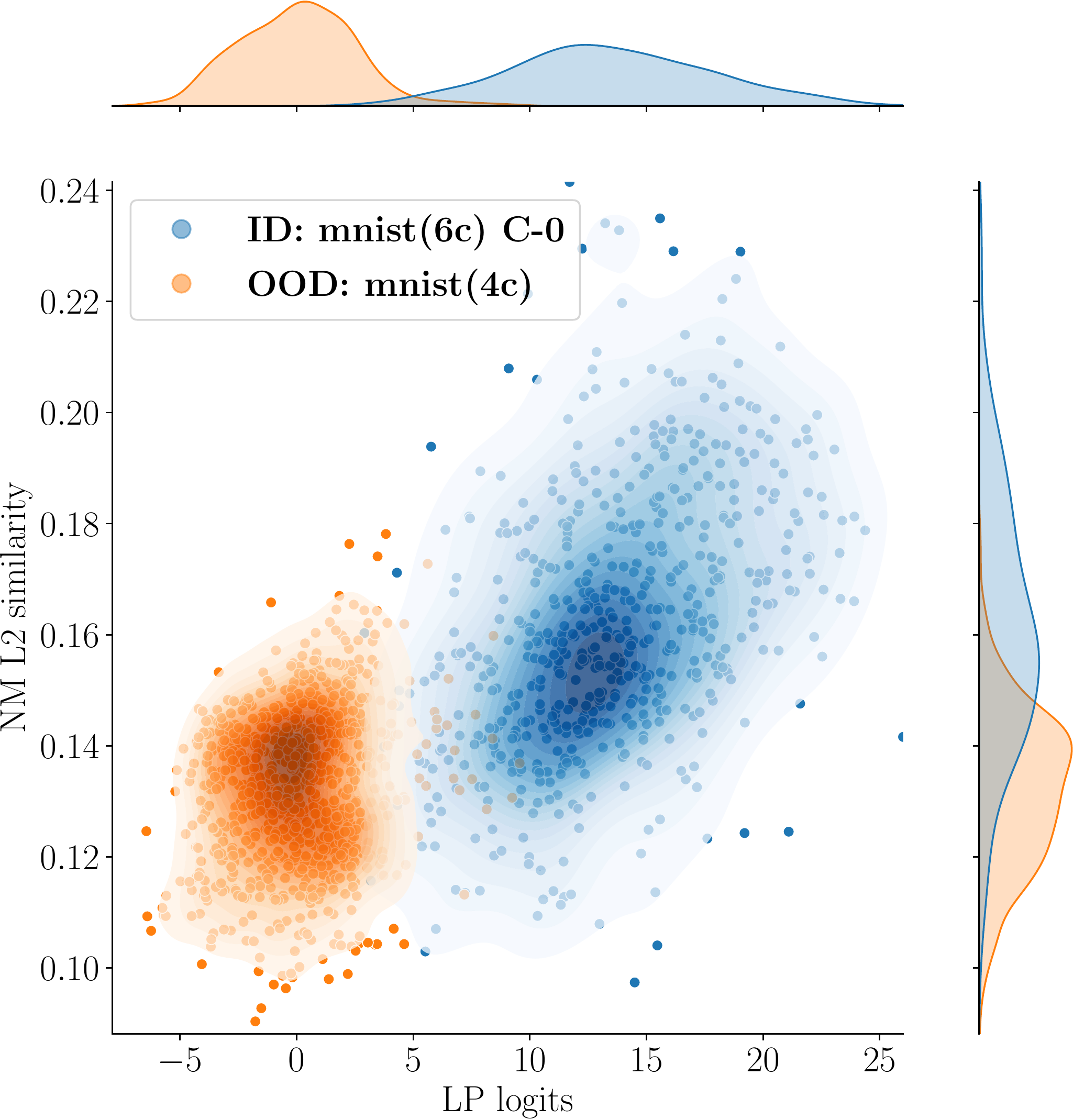} &
	    \includegraphics[width=0.3\linewidth]{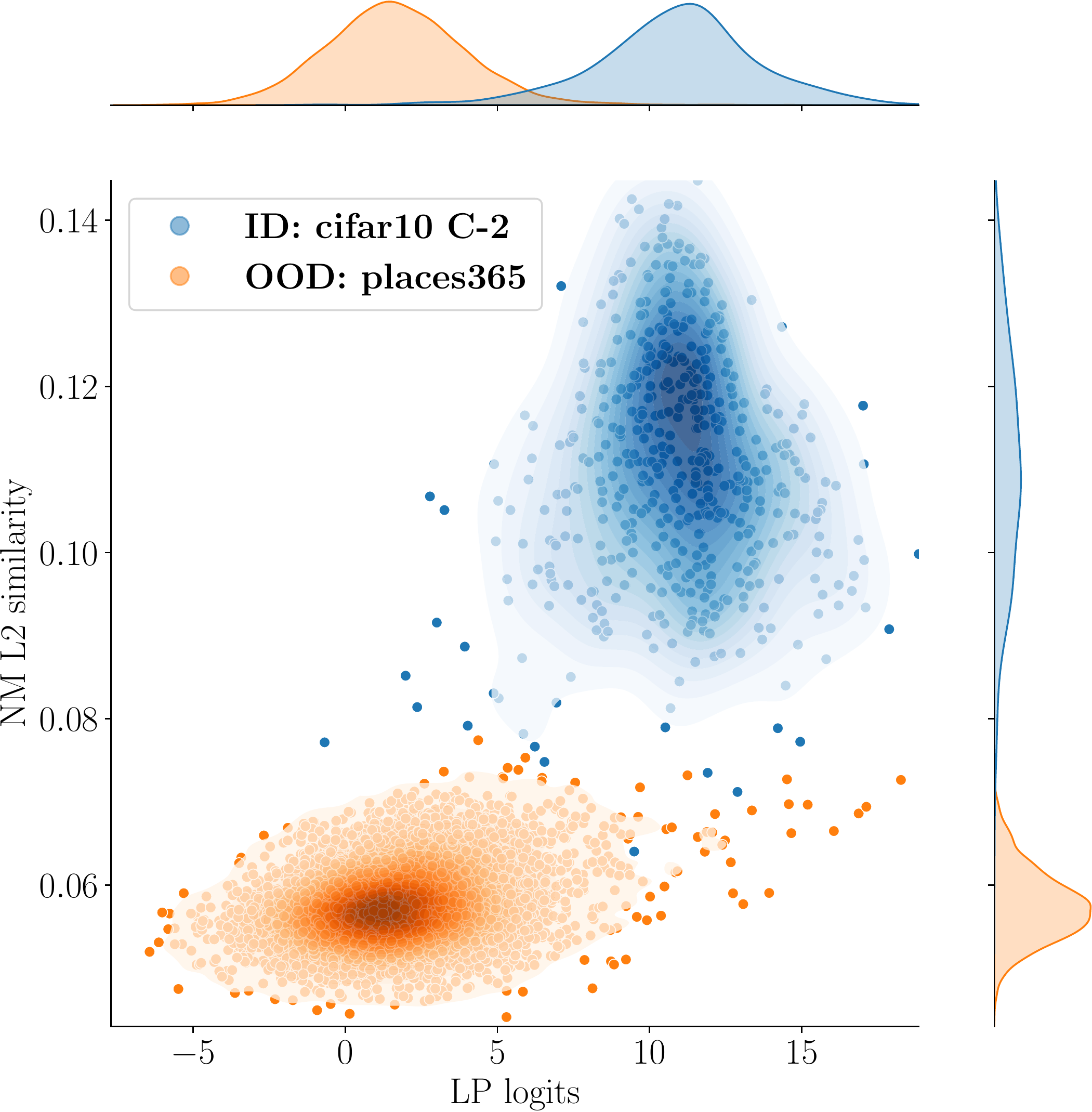} &
	    \includegraphics[width=0.3\linewidth]{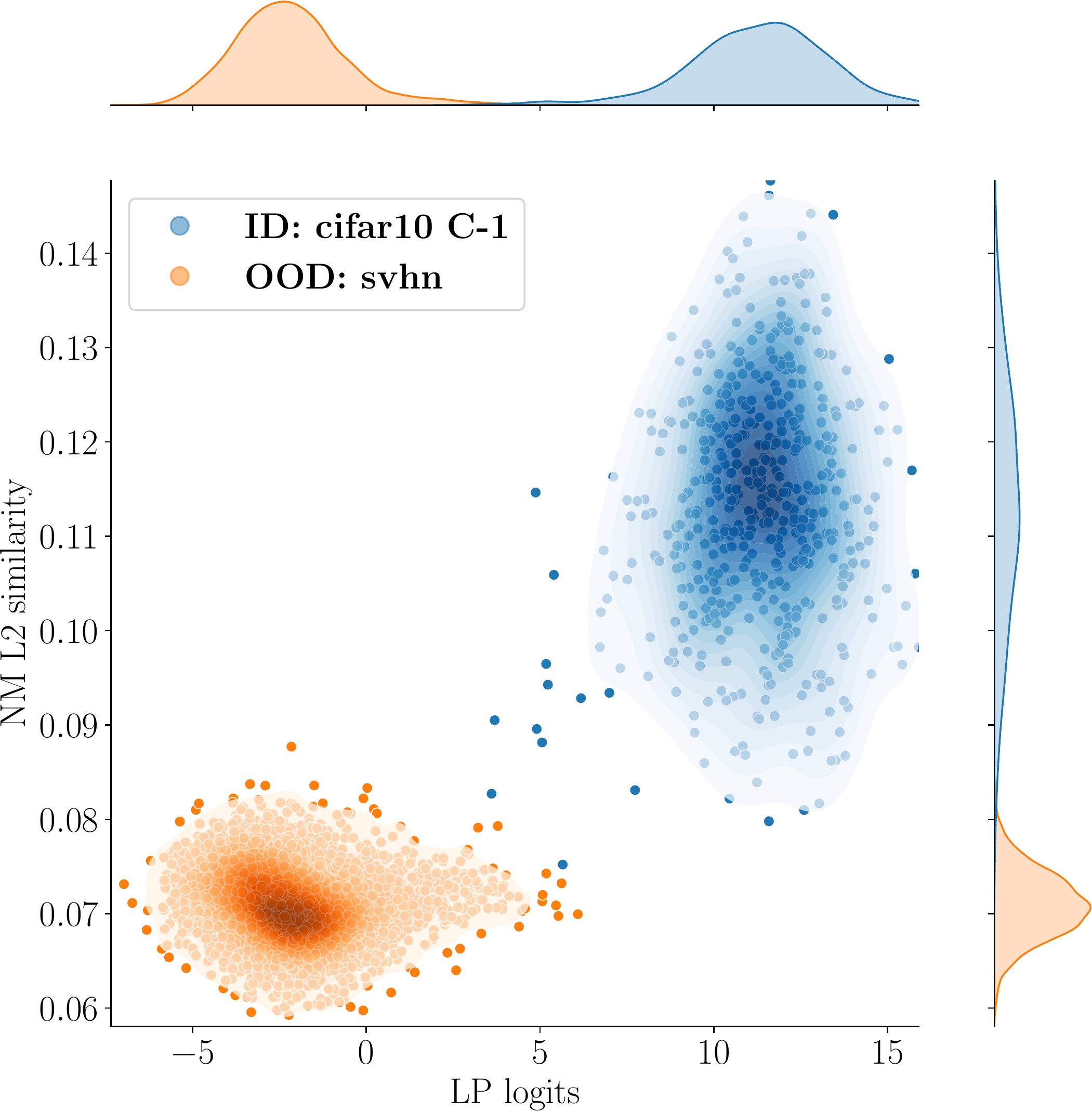} \\ 
	    (a) & (b) & (c) 
    \end{tabular}
    \caption{Complementarity of the LP and NM classifiers. (a) When applied to
    a problem with semantic shift only, the LP classifier tends to separate the
    ID and OOD datasets better. (b) For OOD problems with mixed semantic and
    domain shifts, NM classifier performs typically better. (c) On some
    problems, both perform well. 
    Notice that moving from a single LP or NM similarity to a two-dimensional space
    already allows better separation in {\it all} cases. Compare this with
    detailed results in Tabs~\ref{tab:osr_results}-\ref{tab:semantic_results}.
    }
    \label{fig:lp_ncm_complementarity}
\end{figure*}

\begin{figure}[t]
    \centering
    \includegraphics[width=0.98\columnwidth]{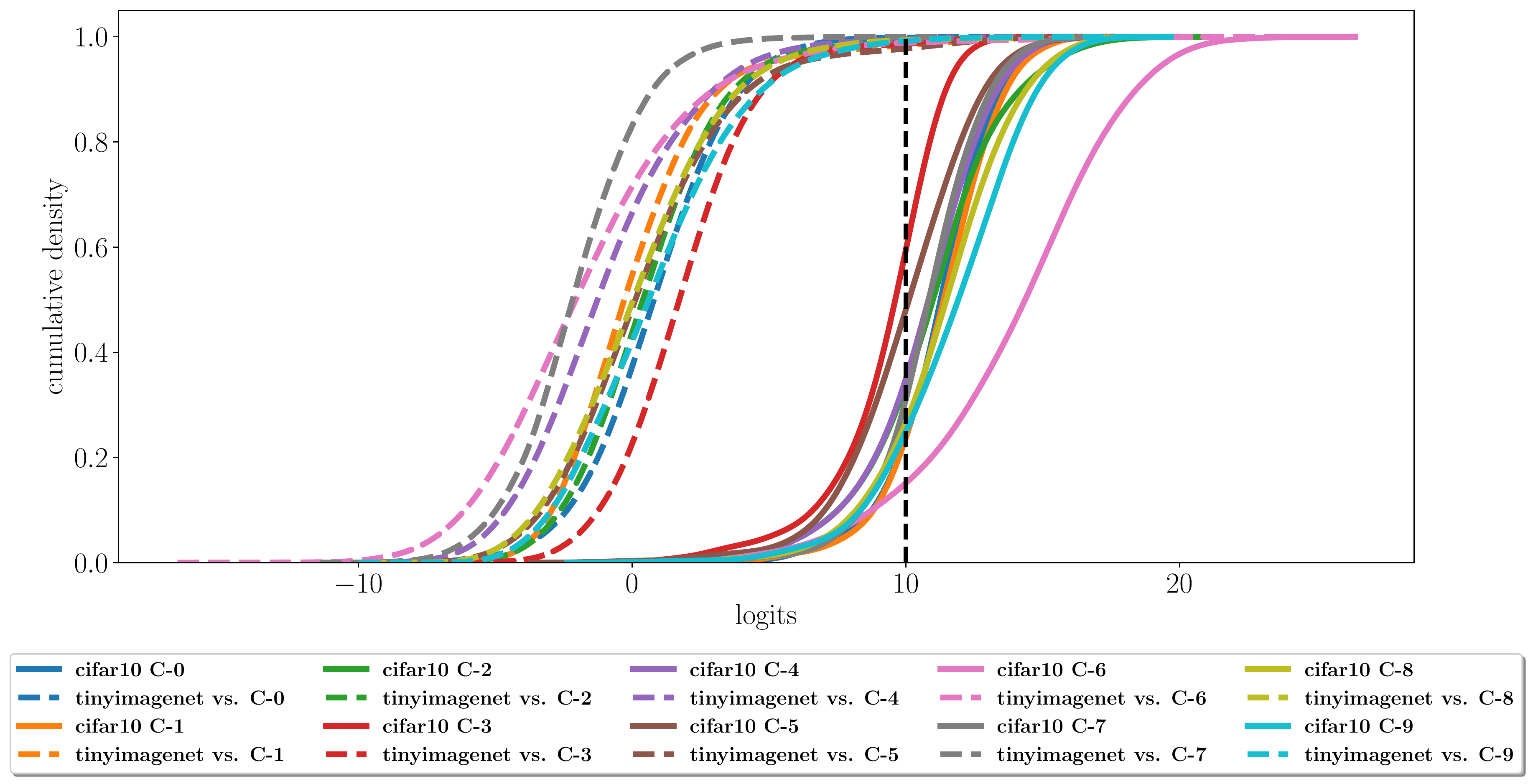}

    \caption{Mis-calibration of the logit scores. The graph shows cumulative
    distributions for ID (full line) and OOD (dashed) classes given the LP
    logit scores trained on the ID data. Here CIFAR10 is ID and TinyImageNet is
    OOD. Selecting a single logit threshold, 10 in this case, results in
    different ID class rejection rates. We call this {\it logits
    mis-calibration}.
    }
    \label{fig:mls_miscalibration}
\end{figure}

\begin{figure}[t]
    \centering
    \includegraphics[width=0.44\textwidth]{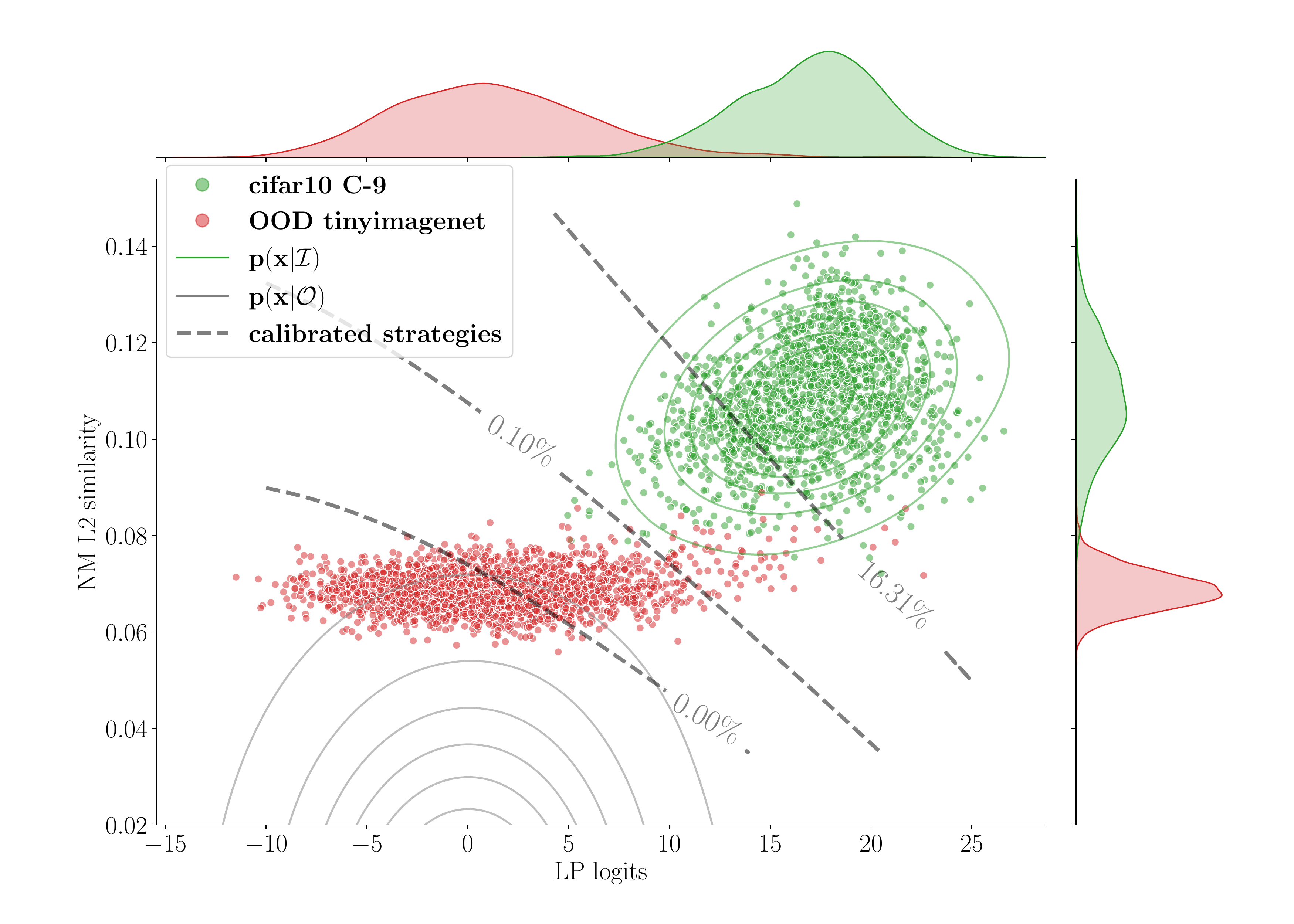}
    \caption{\grood motivation diagram. Class
    9 from CIFAR10 taken as ID and the TinyImageNet dataset as OOD. Two
    classifiers, LP and NM, produce a 2D space to which each sample is mapped to
    (green or red dots for ID and OOD respectively). Top/right axes: the marginal
    empirical distributions. The ID data are modelled
    as a bi-variate Normal distribution (green iso-lines). A "general" OOD distribution is constructed as another Normal distribution (gray
    iso-lines). Three
    possible decision strategies for different expected ID false rejection
    rates in the N-P task are plotted in black dashed lines with corresponding rejection rates
    marked. Note: The proposed methods do not have access to the OOD
    data, they are shown only to strengthen reader's intuition.
    }
    \label{fig:LP_NM}
\end{figure}

In this section we describe the proposed Generic
Representation based OOD detection approach, \grood in short, which exploits a representation
pre-trained on auxiliary large-scale non-OOD-related data. The intuition behind
the method is that a generic representation is
a good starting point for the OOD detection. The proposed method also produces
well-calibrated classification scores for a given ID task corresponding
directly to the same false negative rate for {\it every} ID class.

We expect the representation to be strong, allowing in- and out-of-distribution data
separation by a low-complexity classifier.
In particular, we investigate two such classifiers, Linear Probe (LP) and
Nearest Mean (NM), trained on ID data only. The LP classifier consists of
a single linear projection layer followed by a softmax normalization (\ie
multi-class logistic regression model). This type of classifier has been used
in representation learning to test the expressiveness of
a representation~\cite{Chen2020}. The NM classifier assigns data to
the class with the nearest class mean as measured by the $L_2$ distance;
learning this classifier consists of computation of a mean vector representation
for each class. We chose the LP and NM classifiers because of (i) their
simplicity -- simple classifiers generalize well, do not overfit to ID problem
-- and (ii) complementarity - one is based on a discriminative score and the
other on a distance metric - as illustrated in
Fig~\ref{fig:lp_ncm_complementarity} and Fig~\ref{fig:lp_ncm_complementarity_results}. At the test time, when OOD data points
are detected, the LP and NM classifier responses are simply thresholded (like
in~\cite{Vaze2022}) and this threshold is varied to compute the ROC curves in
the experiments.

Although each of these
classifiers performs already better then state-of-the-art methods on several 
benchmarks, we show in Sec~\ref{sec:complementarity} that they are in fact
complementary, each working better on different type of problems.
Further, as shown in Sec~\ref{fig:mls_miscalibration}, their logit/distance
scores are not well calibrated, \ie when setting an in-out decision threshold,
the ID classes are rejected unevenly, some producing higher false negative
rates then the others.

To solve these issues, we propose a new method, called \grood, which combines
the outputs of the two classifiers. The distribution of the outputs is modelled
as a bi-variate Guassian which permits addressing OOD as a formally defined
two-class Neyman-Pearson task~\cite{Schlesinger2002, Neyman1928, Neyman1933}
through which calibration of the OOD detector is achieved.\footnote{We
experiment with 2D space of scores only, as the amount of data for model
estimation is limited (100 or less examples in some cases).}

We illustrate the approach on an example OOD problem shown in
Fig~\ref{fig:LP_NM}. CIFAR10 is considered ID (class 9 shown here) and the
TinyImageNet represents an OOD dataset (see Sec~\ref{sec:experiments} for
details on datasets). The figure shows in green the ID and in red the OOD
distribution of LP scores (top) and NM similarity (right), see Sec~\ref{sec:LPNM} for the definitions. The
data are plotted as green (ID) and red (OOD) dots. The ID distribution is
specified by the desired ID classification problem, the OOD distribution may
vary depending on particular ID/OOD benchmark. 
Notice, that shifting the problem from a one-dimensional score (either LP or
NM) to a two-dimensional space allows us to leverage the best of LP and NM (cf
Fig~\ref{fig:lp_ncm_complementarity} \&~\ref{fig:lp_ncm_complementarity_results}) and leads to a better ID/OOD separation
when considering jointly {\it all} tested OOD problems.

In the proposed \grood method we model the ID distribution as a Normal
distribution. Although an approximation, we observed empirically that it holds
reasonably well over a wide range of tasks\footnote{A breaking point would be
the case of ID data where one class consists of multiple clusters. In this
case, the NM classifier would need to be modified to consider several
"means".}. Of course, nothing prevents us from using a more complex model of
the distribution, \eg the non-parametric Parzen estimate, if needed, but the
Normal distribution assumption makes the next step in designing \grood
significantly easier.

Next we formulate the ID/OOD classification problem as a multi-class
Neyman-Pearson task~\cite{Schlesinger2002, Neyman1928, Neyman1933}. We start by
considering a single ID class. Let $\mathcal{I}$ be a class representing the ID
samples and $\mathcal{O}$ the class for OOD data. Assume the data are sampled
from a two-dimension domain $\mathcal{X} = \mathcal{X}_{LP} \times
\mathcal{X}_{NM}$, where $\mathcal{X}_{LP}$ is the domain of LP logit scores
and $\mathcal{X}_{NM}$ the domain of NM distances. The task is then to find
a strategy $q^*(x): \mathcal{X} \rightarrow \{\mathcal{I}, \mathcal{O}\}$ such
that
\begin{equation}
\begin{aligned} 
q^* = \argmin_q & \int_{x: q(x) \neq \mathcal{O}} p(x|\mathcal{O})\,dx \\ 
\textrm{s.t.} \quad & \epsilon_\mathcal{I} = \int_{x: q(x) \neq \mathcal{I}} p(x|\mathcal{I})\,dx \leq \epsilon
\end{aligned}
\end{equation}
This optimization problem minimises the false ID
acceptance rate for a particular ID class and bounds the maximal ID rejection
rate by~$\epsilon$. For $K$ classes we specify $K$ such problems and use the
same constant $\epsilon$ for all of them, so that the same fixed rejection rate
is required for all classes.

It is known~\cite{Schlesinger2002} that the optimal strategy for a given
$x\in\mathcal{X}$ is constructed using a the likelihood ratio $r(x)
= p(x|\mathcal{I}) / p(x|\mathcal{O})$ so that $q(x) = \mathcal{I}$ if $r(x)
> \mu$ and $q(x) = \mathcal{O}$ if $r(x) \leq \mu$. The optimal strategy $q^*$
is obtained by selecting the minimal threshold $\mu$ such that
$\epsilon_\mathcal{I} \leq \epsilon$. The problem is solved either analytically
for some simple distributions (such as Gaussian) or numerically otherwise.

To solve this problem we still have to specify the $p(x|\mathcal{O})$
distribution. If we assume this distribution to be uniform in $\mathcal{X}$, we
would decide based on the quantiles of the Normal distribution
$p(x|\mathcal{I})$. However, we constructed $\mathcal{X}$ not from general 1D
random variables, but from the classification scores of LP and NM. It is thus
reasonable to assume that the OOD data will lie in the region where the LP score and NM similarity are low.

To implement this assumption, we construct $p(x|\mathcal{O})$ as another Normal
distribution with a zero mean and a diagonal covariance matrix with large
variances. For the LP, the zero mean assumption is motivated by the
fact that in high dimensional spaces, a random vector is likely to be close to
orthogonal to the id-class directional vectors (the weights of the linear layer
before the softmax). For the NM similarity, the choice was made empirically as 
a limit case for very large $L_2$ distance from a class mean center.
The variance, in both directions, is set so that the range of the data is
a multiple of the in-distribution range, \ie it is very broad. The in-distribution 
range was robustly estimated as a $90\%$ quantile of all in-distribution data scores 
and the multiplicative factor was set empirically. Both Normal distributions for ID and OOD are plotted
in Fig~\ref{fig:LP_NM} in green and gray solid line contours respectively.
Fig~\ref{fig:LP_NM} shows also three optimal strategies (if the assumptions
about normality were true) as gray dashed decision boundaries and their corresponding ID rejection rates. Clearly, the strategy
rejects the least confident ID samples first.

Solving this Neyman-Pearson problem for each ID class gives $K$ strategies
$q^*_k$, each calibrated for the same rejection rate. In practice, we would
specify acceptable rejection rate $\epsilon$ and obtain the optimal strategies
for normally distributed data. For the evaluation where we need to sweep over
the values of $\epsilon$, we sample a limited set of values, find their
likelihood ratio $\mu$ and interpolate in-between.

Finally, for the ID classification we use the $\argmax_{k} p_k(x|\mathcal{I})$ of the 
probabilities obtained as bi-variate Normal distributions for each class $k$.

\section{Experiments}\label{sec:experiments}

In this section we evaluate the proposed \grood method and other
state-of-the-art methods on a wide and diverse set of benchmark problems
described in the OOD literature. We select the benchmarks to cover various
scenarios and to demonstrate the generality of the proposed approach.

\subsection{Benchmarks}

There are several commonly used benchmarks to evaluate OOD and OSR methods and
most papers typically evaluate on their subset. In our evaluation we attempt to
cover most of the commonly used variations. We categorize the experiments based
on the presence/absence of the domain shift (DS) and the semantic shift (SS).

\noindent {\bf No DS, only SS.} For MNIST~\cite{Deng2012},
SVHN~\cite{Netzer2011} and CIFAR10~\cite{Krizhevsky2009} datasets we perform
the 6-vs-4 split~\cite{Chen2021, yangopenood, Vaze2022}. Here six classes are selected as ID at random and the
remaining four as OOD. The experiment is repeated five times with different
splits and the average metrics are reported together with their standard
deviations. 

For a bit larger CIFAR+10 and CIFAR+50 experiments, four classes are sampled
from CIFAR10 and are considered ID and another 10 (or 50) non-overlapping
classes are randomly selected from CIFAR100~\cite{Krizhevsky2009} and used as
OOD~\cite{Chen2021, yangopenood, Vaze2022}. Again, five trials are averaged. For the biggest TIN-20 experiment, twenty
classes are selected randomly as ID and 180 as OOD from the TinyImageNet
dataset~\cite{Torralba2008}. For the above experiments we use the same splits
as in~\cite{Chen2021} for compatibility with previous results.

Finally, to test this type of settings to its limits, we evaluate on the
fine-grained class splits from the Semantic Shift Benchmark~\cite{Vaze2022}.
Here three splits are given: easy, medium and hard, with increasing semantic
shift overlap with ID classes. This overlap is determined from a set of
detailed class attributes. We use the splits for the CUB~\cite{Wah2011}
(birds), StanfordCars~\cite{Krause2013}, and FGVC-Aircraft
datasets~\cite{Maji2013}.

\noindent {\bf DS and SS mixed.} Another common experimental setting is to
consider CIFAR10 as ID and use other datasets as OOD~\cite{liu2020energy, Chen2021, sun2022out, Vaze2022}. In this case there is an
explicit SS and an implicit DS. We evaluate against MNIST~\cite{Deng2012},
SVHN~\cite{Netzer2011}, Textures~\cite{Cimpoi2014}, Places365~\cite{Zhou2017},
CIFAR100~\cite{Krizhevsky2009}, iNaturalist~\cite{Van2018},
TinyImageNet~\cite{Torralba2008} and LSUN datasets~\cite{Xiao2010}.

\noindent {\bf DS only}. A special kind of shift is when the classes stay the
same, but the image domain changes. For this experiment we adopt the benchmark
from~\cite{Chen2021} based on DomainNet dataset~\cite{Peng2019}. The challenge
is to distinguish between photos of objects from 173 classes (ID) and
clipart/quickdraw images from the same classes (OOD). The benchmark also
contains an OOD part with real images from different 173 classes (SS task).

\begin{table}[t]
\caption{
The generic representation models (CLIP, DIHT) consistently outperforms the ImageNet pre-trained
    representation  on a range of OOD tasks. The scores are averages over many
    semantic-shift-only (SS) and mixed SS and domain shift (SS+DS) tasks. The
    SS experiments are the same as in Tab~\ref{tab:osr_results} and SS+DS
    as in Tab~\ref{tab:ood_results}. Evaluation for different network
    architectures and per-dataset results are provided in the supplementary
    material.}
\vspace*{0.3em}
\centering
\resizebox{0.48\textwidth}{!}{
\begin{tabular}{lccS[table-format=2.2]S[table-format=2.2]S[table-format=2.2]S[table-format=2.2]}
\toprule
\multirow{2}{*}{arch} & \multirow{2}{*}{pre-trained} & \multirow{2}{*}{classif} & \multicolumn{2}{c}{SS only} & \multicolumn{2}{c}{SS + DS} \\
\cmidrule(lr){4-5} \cmidrule(lr){6-7}
 &  &  & \multicolumn{1}{c}{AUROC$\uparrow$} & \multicolumn{1}{c}{OSCR$\uparrow$} &  \multicolumn{1}{c}{AUROC$\uparrow$} & \multicolumn{1}{c}{FPR95$\downarrow$} \\
\midrule
    \multirow{2}{*}{ViT-L/16}     & ImageNet & LP & 91.82           & 86.92           & 95.98           & 21.20 \\
                                  & ImageNet & NM & 77.67           & 69.05           & 81.48           & 69.34 \\
    \multirow{2}{*}{ViT-L/14}     & DIHT     & LP & 93.62 & 89.91&\first{99.27} & \first{\pz2.68}\\
                                  & DIHT     & NM & 88.44 & 81.80 & 99.15 & \pz4.26\\
    \multirow{2}{*}{ViT-L/14}     & CLIP     & LP & \first{94.35} & \first{91.10}     & 97.01           & \pz8.73 \\
                                  & CLIP     & NM & 85.05           & 79.26           & 98.06 & \pz8.62 \\
\bottomrule
\end{tabular}
}

\label{tab:repr_comparison}
\end{table}

\begin{figure*}[t]
    \centering
    \begin{tabular}{cc}
	    \includegraphics[width=0.48\linewidth]{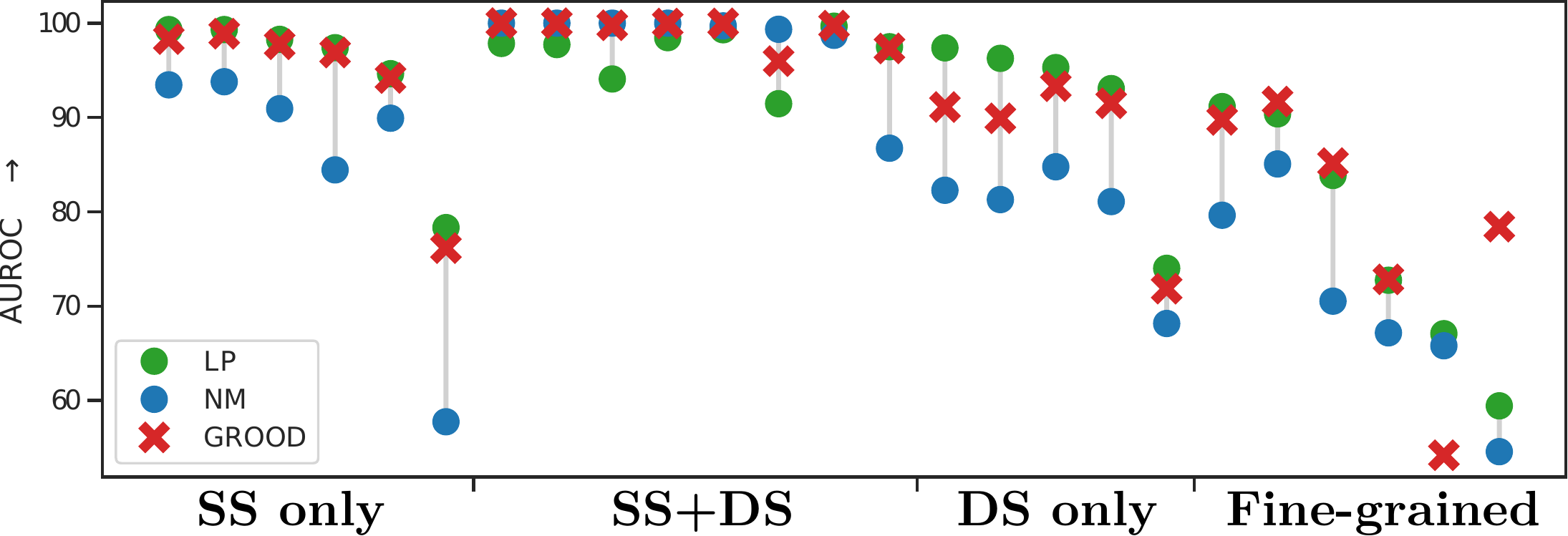} &
	    \includegraphics[width=0.48\linewidth]{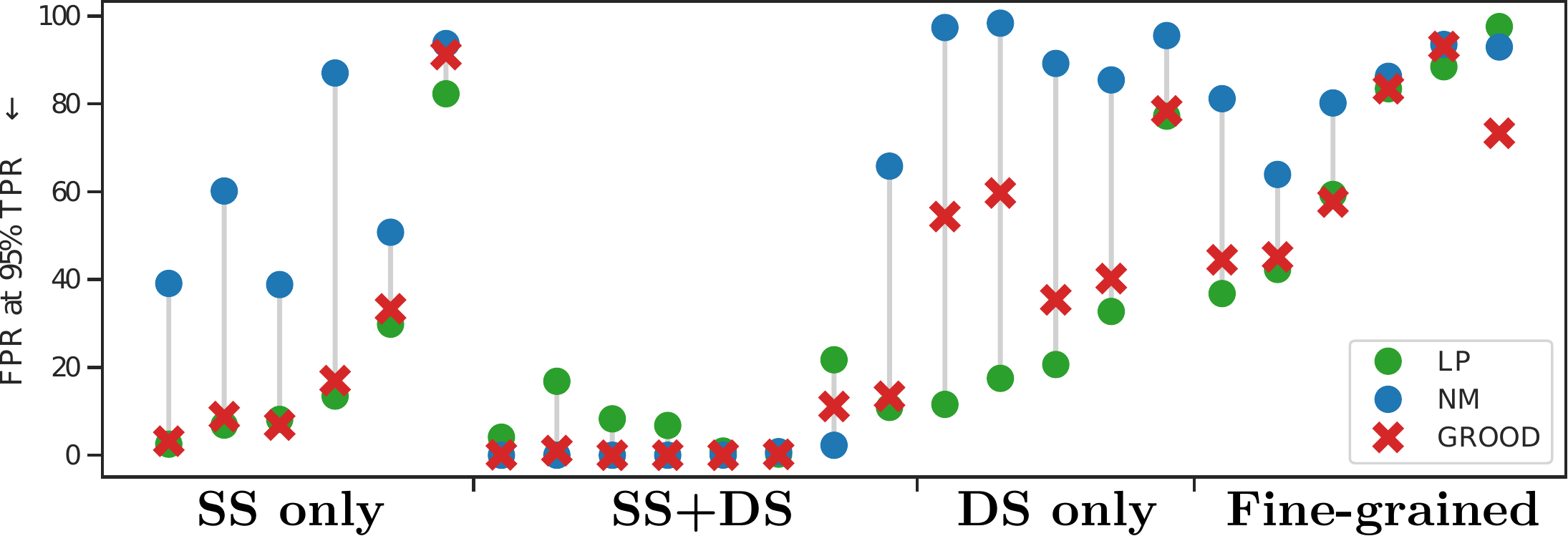} \\ 
	    (a) & (b)
    \end{tabular}
    \caption{Results of LP and NM classifiers and the proposed \grood method on various OOD tasks. Every datapoint on x-axis corresponds to one type of experiment (\eg CIFAR10 vs. MNIST). The figure illustrates the complementarity of LP and NN and shows that the proposed \grood method is better choice without a priori knowledge about the type of OOD data. The results show (a) AUROC and (b) FPR at 95\% TPR metric. SS and DS refers to semantic and distribution shift respectively.}
    \label{fig:lp_ncm_complementarity_results}
\end{figure*}

\begin{table*}[t]
\caption{Comparison with the state-of-the-art -- OOD problems with semantic shift only. For the description of the measures see Sec~\ref{sec:metrics}.}
\vspace*{0.3em}
\centering\tiny\resizebox{0.85\textwidth}{!}{
\begin{tabular}{llcccccc}
\toprule
{} & {} & \multicolumn{6}{c}{AUROC $\uparrow$} \\

\cmidrule(lr){3-8}

{} & {from}  &
\multicolumn{1}{c}{MNIST} & 
\multicolumn{1}{c}{SVHN} & 
\multicolumn{1}{c}{CIFAR10} & 
\multicolumn{1}{c}{CIFAR+10} & 
\multicolumn{1}{c}{CIFAR+50} & 
\multicolumn{1}{c}{TIN}  \\ 

\midrule

Deep KNN~\cite{sun2022out}                     & \cite{yangopenood} & 97.50         & \na          & 86.90         & \na          & \na          & 74.10   \\
DeepEnsemble~\cite{lakshminarayanan2017simple} & \cite{yangopenood} & 97.20         & \na          & 87.80         & \na          & \na          & 76.00  \\
Pixmix~\cite{hendrycks2022pixmix}              & \cite{yangopenood} & 93.90         & \na          & 90.90         & \na          & \na          & 73.50   \\

\cmidrule(lr){1-8}

OpenHybrid~\cite{Zhang2020HybridMF}            & \cite{Vaze2022}    & 99.50 & 94.70         & 95.00         & 96.20         & 95.50         & 79.30  \\
MLS~\cite{Vaze2022}                            & \cite{Vaze2022}    & 99.30          & \first{97.10}         & 93.60         & 97.90         & 96.50         & 83.00  \\

ARPL+CS~\cite{Chen2021}                        & \cite{Chen2021}    & \first{99.70$\pm$0.10} & 96.70$\pm$0.20 & 91.00$\pm$0.70 & 97.10$\pm$0.30 & 95.10$\pm$0.20 & 78.20$\pm$1.30 \\

{\bf \grood}                                         & ours & 96.88$\pm$0.47 & 76.14$\pm$1.14 & \first{97.76$\pm$0.43} & \first{98.88$\pm$0.38}         & \first{98.31$\pm$0.25}          & \first{94.18$\pm$0.94}  \\

\bottomrule
\end{tabular}
}
\label{tab:osr_results}
\end{table*}

\subsection{Evaluation Metrics}
\label{sec:metrics}

There seems to be no consensus, which metrics to report for OOD detection. The
most commonly used is the AUROC metric, which measures the ability to
distinguish OOD data from ID data. Often this is the only metric reported even
though it does not show, how well the method performs on the ID classification
task. For ID classification people report either the ID accuracy, FPR95 or OSCR
score. In the tables in Sec~\ref{sec:experiments} we report the most commonly
used metric for particular OOD problem.

Assuming a binary ID vs OOD classification problem, the {\bf AUROC} measures
the area under the true positive (TP) -- false positive (FP) rates curve, where
the ID data is considered be the positive class. We adopt the evaluation code
from~\cite{Vaze2022, Chen2021}.

The {\bf FPR at $95\%$ TPR (FPR95)} metric measures the false positive rate at
$95\%$ true positive rate on the same binary problem as the AUROC measure.

The Open-Set Classification Rate {\bf (OSCR)}~\cite{Dhamija2018oscr, Chen2021}
measures the trade-off between the ID classification accuracy and OOD detection
accuracy. It is computed as area under
$\mathrm{CCR}(\theta)$-$\mathrm{FPR}(\theta)$ curve where
$\mathrm{CCR}(\theta)$ is the correct classification rate defined as
\begin{equation}
    \mathrm{CCR}(\theta) =
    \frac{|\{x\in \mathcal{T}_k | \argmax_j p(j|x) = k \land p(k|x) \ge \theta\}|}{|\mathcal{T}_k|}\,,
    \end{equation}
    where $\mathcal{T}_k$ is the sub-set of the ID training data belonging to the class $k$, and $\mathrm{FPR}(\theta)$ is the false positive rate defined as
\begin{equation}
    \mathrm{FPR}(\theta) = \frac{|\{x \in \mathcal{U} | \max_k p(k|x) \ge \theta \}|}{|\mathcal{U}|}\,,
\end{equation}
where $\mathcal{U}$ is the set of OOD data available at the test time.

Finally, the {\bf ACC} measures the accuracy on the ID classification problem.

\begin{table*}[t]
\caption{Comparison with the state-of-the-art --  OOD problems with mixed semantic and domain shifts. The method marked with $\ddagger$ was trained by us. The measures are described in Sec~\ref{sec:metrics}.}
\vspace*{0.3em}
\centering\tiny\resizebox{\textwidth}{!}{
\begin{tabular}{lccccccccc}
\toprule
{} & {} & \multicolumn{8}{c}{FPR at $95\%$ TPR $\downarrow$ $\biggm/$ AUROC $\uparrow$} \\

\cmidrule(lr){3-10}

{} & from &
\multicolumn{1}{c}{SVHN} & 
\multicolumn{1}{c}{MNIST} & 
\multicolumn{1}{c}{Textures} & 
\multicolumn{1}{c}{Places365} & 
\multicolumn{1}{c}{CIFAR-100} & 
\multicolumn{1}{c}{iNaturalist} & 
\multicolumn{1}{c}{TIN} & 
\multicolumn{1}{c}{LSUN} \\

\midrule

\multirow{2}{*}{{Deep KNN~\cite{sun2022out}}}  & \cite{yangopenood} & 33.32 / 95.13   & 50.08 / 91.63   & 46.01 / 92.77   & 43.78 / 91.82 & 52.49 / 89.55 & --- & 46.66 / 91.41 & ---  \\
            {}                                 & \cite{sun2022out}  & \pz2.40 / 99.52 & ---             & \pz8.09 / 98.56 & 23.02 / 95.36 & ---           & --- & ---           & \pz1.78 / 99.48 \\
LogitNorm~\cite{wei2022mitigating}             & \cite{yangopenood} & \pz5.30 / 98.86 & \pz4.75 / 98.82 & 30.94 / 94.30   & 31.17 / 94.76 & 46.99 / 91.13 & --- & 36.34 / 93.90 & ---        \\
UDG~\cite{Yang2021SemanticallyCO}              & \cite{yangopenood} & 61.91 / 92.50   & 39.32 / 93.81   & 43.97 / 93.56   & 42.44 / 93.58 & 55.33 / 90.38 & --- & 42.48 / 93.33 & ---         \\
DeepEnsemble~\cite{lakshminarayanan2017simple} & \cite{yangopenood} & 37.03 / 94.95   & 41.65 / 94.34   & 48.39 / 92.59   & 50.20 / 91.06 & 54.31 / 89.76 & --- & 48.93 / 91.35 & ---          \\
Pixmix~\cite{hendrycks2022pixmix}              & \cite{yangopenood} & 13.70 / 98.01   & 49.72 / 91.78   & \pz8.07 / 98.83 & 38.51 / 94.03 & 47.12 / 91.81 & --- & 36.47 / 94.31 & ---           \\

ARPL+CS~\cite{Chen2021} & \pz$\ddagger$ & 53.20 / 90.64  & 42.44 / 94.13  & 51.84 / 90.68&47.42 / 90.72 & 57.11 / 88.52 & 56.02 / 89.73 & 53.40 / 88.61 & 46.32 / 91.85 \\

SSD~\cite{sehwag2021ssd} & \cite{sehwag2021ssd} &  \hspace{0.5em} --- / 99.6 & --- & \hspace{0.5em} --- / 97.6 & \hspace{0.5em} --- / 95.2 & \hspace{0.5em} --- / 90.6 & --- & --- & \hspace{0.5em} --- / 96.5 \\


    {\bf \grood} & ours & \first{\pz0.00 / 99.97}  & \first{\pz0.20 / 99.74}  & \first{\pz0.09 / 99.96} & \first{\pz1.05 / 99.78} & \first{13.41 / 97.32} & \first{\pz0.00 / 100.00} & \first{11.11 / 95.97}& \first{\pz0.00 / 100.00} \\

\bottomrule
\end{tabular}
}

\label{tab:ood_results}
\end{table*}

\begin{table*}[t]
\caption{Comparison with the state-of-the-art -- OOD problems with a semantic shift (Real-B column) and domain shift only (next four columns). For the description of the measures see Sec~\ref{sec:metrics}.}
\vspace*{0.3em}
\centering\tiny\resizebox{0.8\textwidth}{!}{
        \begin{tabular}{llccccc}
            \toprule
            {} & {} & \multicolumn{5}{c}{ID: Real-A$(0\dots 172)$ $\;\;\left[\right.$AUROC $\uparrow$ / OSCR $\uparrow$$\left.\right]$} \\

            \cmidrule(lr){3-7}

            {} & {from} &
            \multicolumn{1}{c}{Real-B} & 
            \multicolumn{1}{c}{Clipart-A} & 
            \multicolumn{1}{c}{Clipart-B} & 
            \multicolumn{1}{c}{Quickdraw-A} & 
            \multicolumn{1}{c}{Quickdraw-B} \\

            \midrule

            ARPL+CS  &   \cite{Chen2021} & 75.20 / 61.90 & \first{72.70} / 59.40 &  82.90 / 66.60 & 86.70 / 69.00 &  87.50 / 69.50 \\


            {\bf\grood} & ours & \first{91.50 / 83.54} & 71.84 / \first{65.20} & \first{93.31 / 85.19} & \first{89.91 / 81.98}& \first{91.11 / 82.83} \\

            \bottomrule
        \end{tabular}
   }
    \label{tab:clipart_results}
\end{table*}

\subsection{Low-Complexity Classifiers}
\label{sec:LPNM}

As we argue for a good and general enough representation as the basis for the
OOD detection, we use only simple low-complexity classifiers (\ie letting the
representation play the essential part in the decision). In particular, we use
the Linear Probe (LP) and the Nearest Mean (NM) classifiers. The LP classifier is trained on the ID data only. We use the training code
from~\cite{radford2021learning}. As an OOD detection score we use the Maximum
logit~\cite{Vaze2022}. The NM classifier's means are also estimated on the ID data only. The NM 
similarity is computed from the NM $L_2$ distance $d_{NM}$ as $1 / (1 + d_{NM})$.

\subsection{Power of a Good Representation}
We start by investigating the effect of different pre-trained representations
on various OOD problems. We consider three rich representations: one trained with
full supervision on the ImageNet1k classification task, and two self-supervised representations, CLIP~\cite{radford2021learning} and DIHT~\cite{Radenovic_2023_CVPR}.

The ImageNet pre-trained representation proved to be a strong baseline for many
problems in computer vision. We use the ViT-L/16 model pre-trained on
ImageNet1k available in the PyTorch Torchvision library and use its penultimate
layer as a feature extractor. It produces 1048-dimensional feature vectors.

The CLIP/DIHT representations have shown outstanding performance on various zero-shot
classification problems~\cite{radford2021learning, Radenovic_2023_CVPR} demonstrating their
versatility. From the point of view of OOD detection, what makes the
representation appealing is that it was trained on image-text pairs instead of
a fixed set of classes. This, together with the self-supervised training
possibly allows the model to extract very rich representation of the visual
world. This makes it a good candidate for separating ID classes from OOD data
irrespective of the type of semantic and distribution shift if these shifts are
covered by natural language and represented sufficiently by the training data.
The CLIP and DIHT models (we are using only the image encoder) produce 768-dimensional
feature vectors.

We have also considered smaller ImageNet and CLIP models, but they perform consistently worse, please refer to supplementary materials for smaller models results.

For all representations, we train the LP and NM classifiers and test over
a range of tasks. We obtain consistent relative performance over different OOD
tasks hence we report only the average metrics. We refer to supplementary
materials for full results.  Tab~\ref{tab:repr_comparison} reports the
average performance on the studied benchmarks.  Our results show that self-supervised
representation works better irrespective of the classifier and the type of OOD
shift. We chose the CLIP representation in the following experiments.

\subsection{Complementarity of LP and NM}
\label{sec:complementarity}

Fig~\ref{fig:lp_ncm_complementarity_results} shows that the LP and NM classifiers are
complementary, each performing well on different types of OOD data. 
This observation motivated the proposed method. Without a priori information, \grood is able to use efficiently the information provided by both LP and NM and in most cases achieve performance of the better one. 

Fig~\ref{fig:lp_ncm_complementarity} further illustrates 
LP and NM scores distributions for different ID and OOD datasets.
We observed that for OOD tasks where ID and OOD classes are from the same
domain (\eg~6-vs-4 experiment on MNIST) and are thus close to each other in the
considered representation, LP tends to work better by finding a suitable linear
projection where the ID classes can be well separated whereas the NM classifier
struggles distinguishing small distances in the high-dimensional representation
(Fig~\ref{fig:lp_ncm_complementarity}a). When the ID and OOD classes are from
rather distant domains (\eg~CIFAR10 and Places365), the NM method works better
as the $L_2$ distance starts to be discriminative enough
(Fig~\ref{fig:lp_ncm_complementarity}b). And there are some problems (\eg~CIFAR10 vs SVHN) where both classifiers produce similarly good separation
between ID and OOD classes (Fig~\ref{fig:lp_ncm_complementarity}c).

\subsection{Mis-calibration of the Logit Scores}
\label{sec:miscalibration}

Another issue revealed in our experiments is mis-calibration of the maximum
logit (or probability) approaches~\cite{Vaze2022,hendrycks2016baseline,liang2017enhancing,hendrycks2019anomalyseg}. We demonstrate this in
Fig~\ref{fig:mls_miscalibration}. When a logit score threshold is
selected (10 in the figure), it produces different false negative (FN)
rates for each class. This is in contrast with \grood method, where the
threshold is imposed directly on the class FN rate. This
allows to specify an allowed FN rate while minimizing the FP rate (\ie the
number of OOD data classified as ID). This quality is important in
safe-critical applications where certain classes are reported as OOD more often
or in social-related applications where having uneven FN rates on ID classes
may lead to unwanted biases.

\subsection{\grood vs State-of-the-Art}

\begin{table*}[t]
\caption{Comparison with the state-of-the-art -- OOD problems with fine-grained semantic shift.
Particularly difficult cases, included to highlight the limitations of the CLIP pre-trained representation (possible future work).
For the description of the measures see Sec~\ref{sec:metrics}.}
\vspace*{0.25em}
\centering\tiny\resizebox{\textwidth}{!}{
        \begin{tabular}{llccccccccc}
            \toprule
            {} & {} & \multicolumn{3}{c}{CUB [Easy / Hard]} & \multicolumn{3}{c}{SCars [Easy / Hard]} & \multicolumn{3}{c}{FGVC-Aircraft [Easy / Hard]} \\ 

            \cmidrule(lr){3-5}
            \cmidrule(lr){6-8}
            \cmidrule(lr){9-11}

            {} & from &
            \multicolumn{1}{c}{ACC} & 
            \multicolumn{1}{c}{AUROC} & 
            \multicolumn{1}{c}{OSCR} & 
            \multicolumn{1}{c}{ACC} & 
            \multicolumn{1}{c}{AUROC} & 
            \multicolumn{1}{c}{OSCR} & 
            \multicolumn{1}{c}{ACC} & 
            \multicolumn{1}{c}{AUROC} & 
            \multicolumn{1}{c}{OSCR} \\ 

            \midrule
            ARPL+ & \cite{Vaze2022} & 85.90 & 83.50 / 75.50 & 76.00 / 69.60 & 96.90          & \first{94.80} / 83.60  & \first{92.80} / 82.30  & 91.50 & 87.00 / 77.70 & 83.30 / 74.90 \\
MLS   & \cite{Vaze2022} & 86.20 & 88.30 / \first{79.30}  & 79.80 / \first{73.10}  & \first{97.10} & 94.00 / 82.20 & 92.20 / 81.10 & \first{91.70}  & \first{90.70 / 82.30}  & \first{86.80 / 79.80} \\


            {\bf\grood} & ours & \first{90.12}          & \first{91.69} / 72.83  & \first{82.49} / 65.38 & 96.82         & 89.74 / \first{85.16}  & 86.79 / \first{82.31} & 70.8  & 78.42 / 54.18 & 58.40 / 42.62 \\

            \bottomrule
        \end{tabular}
    }
    \label{tab:semantic_results}
\end{table*}

Finally, we compare \grood with state-of-the-art methods on an extensive range
of benchmarks. See the results in
Tab~\ref{tab:osr_results}-\ref{tab:semantic_results}. In all the tables we
compare against a selection of best performing methods collected from
literature and indicate the respective source publication. For comparison with
many other methods reported earlier and with worse results see the benchmark papers~\cite{yangopenood, yang2021generalized, Vaze2022}.

Tab~\ref{tab:osr_results} and Tab~\ref{tab:ood_results} summarize the most
common benchmarks used in literature, the first one with the semantic shift
only and the other with mixed semantic and domain shifts. 
Compared to state-of-the-art methods, \grood outperforms all of them by a large
margin on most of the OOD problems. Especially on the mixed 
semantic and domain shift 
problems in Tab~\ref{tab:ood_results}, our approach basically solves all
the benchmarks.

The proposed method is the most effective on more complex problems like CIFAR
variants and TIN and struggles a bit on the 6-vs-4 SVHN problem in
Tab~\ref{tab:osr_results}. We attribute this mainly to the dataset ground truth
construction. The images do not contain the single digit stated in the GT label but
also ``some distracting digits to the sides of the digit of
interest''\footnote{\url{http://ufldl.stanford.edu/housenumbers}}). For CLIP
which was trained on many images containing text (with full text label) all
digits in the image influence the representation. The performance drop does not
happen for MNIST dataset with a single digit per-image, which supports our
analysis. Since methods that train the representation on ID data do not suffer
from this phenomena, these issues can be potentially alleviated by fine-tuning
the representation or using more complex classifiers.

\grood is also very efficient on the problems with domain shift only as shown
in Tab~\ref{tab:clipart_results}. Here our method again outperforms the current
state-of-the-art significantly, showing the ability to distinguish data even
along such distribution shifts like real-image vs clipart vs quick draw.
There is still a space for improvement on the Clipart-A split which is very
similar to the ID Real-A dataset (same classes, photos vs complex clipart).
This is though difficult even for ARPL+CS which is trained on the ID data.

Finally, to test the limits of the proposed method we evaluated \grood on the
Semantic Shift Benchmark problems with very fine-grained semantic shift in
Tab~\ref{tab:semantic_results}. Although the class separation is often very
subtle, our method performs comparably to state-of-the-art. 
The Easy/Hard splits in SCars and FGVC-Aircrafts datasets (in contrast
to CUB) are not based on strictly visual attributes, but on attributes like
the year of production or an aircraft variant. They do not seem to correspond to
differences captured by the CLIP representation. This is more pronounced in
case of the airplanes where for instance the overall shape do not change between
production years (as oppose to cars). We see this as a border case and
a weakness of the benchmark\footnote{``...open-set bins of different difficulties in
Stanford Cars are the most troublesome to define. This is because the rough
hierarchy in the class names may not always correspond to the degree of visual
similarity between the classes''\cite{Vaze2022}.} and a possible future direction of research.

Overall, the experiments demonstrate how using a powerful representation leads
not only to a state-of-the-art ID classification as demonstrated
earlier~\cite{radford2021learning}, but provides a classifier with very
strong cues for OOD detection as well.

\section{Conclusions}

In this paper we propose a novel approach to OOD detection which uses a generic
pre-trained representation instead of training a discriminative classifier on
the ID classes.

We model the classification scores of the LP and NM classifiers
for the ID classes as a multivariate Guassian and show that this permits
addressing OOD detection as a formally defined two-class Neyman-Pearson
task. Compared to traditional logit (or distance) thresholding, the solution to this task
leads to naturally calibrated OOD detection score connected directly to the
same false negative rate on all ID classes. Moreover, the resulting \grood
method leverages the strengths of both used classifiers leading to
consistent performance over all considered benchmarks.

The proposed \grood method was compared to the state-of-the-art methods on
a wide and diverse range of OOD problems with various types and strengths of semantic
and domain shifts. It effectively solves the mixed semantic and distribution
shift benchmarks and achieves the best performance on most of the other
considered problems.

The only observed limitations are related to the interaction of the annotation in
SVHN dataset with the CLIP representation, and very fine-grained classification of airplanes which goes beyond visual attributes.

The simplicity of the adaptation of the \grood method to a novel problem --
only a multi-class logistic regression and finding the mean of each class is needed for training -- makes the process fast.

We suggest that the proposed method combined with a generic representation is
suitable for most OOD tasks based on natural images;
with \grood many of the standard benchmarks are saturated and no longer
stimulate further progress. A recently introduced NINCO dataset~\cite{bitterwolf2023ninco} may be a possible direction, but is left as a future work.

\noindent{\bf Acknowledgement}
This work was supported by Toyota Motor Europe. We would also like to thank
Jonáš Šerých for many valuable discussions related to this paper.

{\small
\bibliographystyle{ieee_fullname}
\bibliography{pood}
}

\end{document}